\documentclass[letterpaper, 10 pt, conference]{ieeeconf}  

\IEEEoverridecommandlockouts                              

\usepackage{graphics} 
\usepackage{epsfig} 
\usepackage{mathptmx} 
\usepackage{times} 
\usepackage{amsmath} 
\usepackage{amssymb}  

\usepackage{hyperref}
\usepackage{optidef}
\usepackage{multirow}
\usepackage{todonotes}
\usepackage{siunitx}
\usepackage{subcaption}
\usepackage{float}
\usepackage{stfloats}

\renewcommand{\baselinestretch}{0.985} 

\newtheorem{remark}{Remark}

\title{\LARGE \bf
Towards Using Fast Embedded Model Predictive Control for Human-Aware Predictive Robot Navigation
}

\author{Till Hielscher$^{1}$,  Lukas Heuer$^{2}$, Frederik Wulle$^{3}$, Luigi Palmieri$^{2}$ 
	\thanks{$^{1}$ T. Hielscher is with the Socially Intelligent Robotics Lab, Institute for Artificial Intelligence, University of Stuttgart, Germany, \texttt{till.hielscher@ki.uni-stuttgart.de}}%
	\thanks{$^{2}$ L. Heuer, L. Palmieri are with Bosch Corporate Research, Stuttgart, Germany, \texttt{lukas.heuer, luigi.palmieri@de.bosch.com}}%
  	\thanks{$^{3}$ F. Wulle is with ARENA2036 e.V., a research campus at the University of Stuttgart, Germany, \texttt{ frederik.wulle@arena2036.de}.}%
   \thanks{This work was partly supported by the EU Horizon 2020 research and innovation program under grant agreement No. 101017274 (DARKO).}%
}

\begin{document}
\maketitle
\thispagestyle{empty}
\pagestyle{empty}
\begin{abstract}
    Predictive planning is a key capability for robots to efficiently and safely navigate populated environments. Particularly in densely crowded scenes, with uncertain human motion predictions, predictive path planning, and control can become expensive to compute in real time due to the curse of dimensionality. With the goal of achieving pro-active and legible robot motion in shared environments, in this paper we present \texttt{HuMAN-MPC}, a computationally efficient algorithm for \texttt{Hu}man \texttt{M}otion \texttt{A}ware \texttt{N}avigation using fast embedded \texttt{M}odel \texttt{P}redictive \texttt{C}ontrol. The approach consists of a novel model predictive control (MPC) formulation that leverages a fast state-of-the-art optimization backend based on a sequential quadratic programming real-time iteration scheme while also providing feasibility monitoring.
    Our experiments, in simulation and on a fully integrated ROS-based platform, show that the approach achieves great scalability with fast computation times without penalizing path quality and efficiency of the resulting avoidance behavior.
\end{abstract}
\section{INTRODUCTION}
    The usage of autonomous mobile robots (AMRs) is continuously increasing in various types of environments for accomplishing several tasks, e.g. pick and delivery in logistic centers, and cleaning of cluttered surfaces. In these environments, robots share spaces with humans.
    To accomplish this the robot needs to infer how surrounding humans will move in the near future and utilize those predictions in its navigation framework \cite{rudenkoIJRR2020}. Moreover, to respond to dynamic changes in surrounding environments, robots need to compute actions in real time. Model predictive control (MPC) is an established method to compute robot trajectories in such settings \cite{schoels2020ciao, heuerIROS2023}. However, those techniques suffer from the curse of dimensionality in such environments, i.e., planning considering human motion predictions is quite expensive \cite{schoels2020ciao, heuerIROS2023, schoels2020nmpc, schaefer2021leveraging}.
    
    With the goal of handling the introduced complexity and achieving fast and reliable robot motion generation in crowded environments, we present \texttt{HuMAN-MPC} -- a system for \texttt{Hu}man \texttt{M}otion \texttt{A}ware \texttt{N}avigation using fast embedded \texttt{M}odel \texttt{P}redictive \texttt{C}ontrol and make the following contributions:
    \begin{figure}[h]
        \vspace{-15pt}
            \centering
            \includegraphics[width=0.35\textwidth]{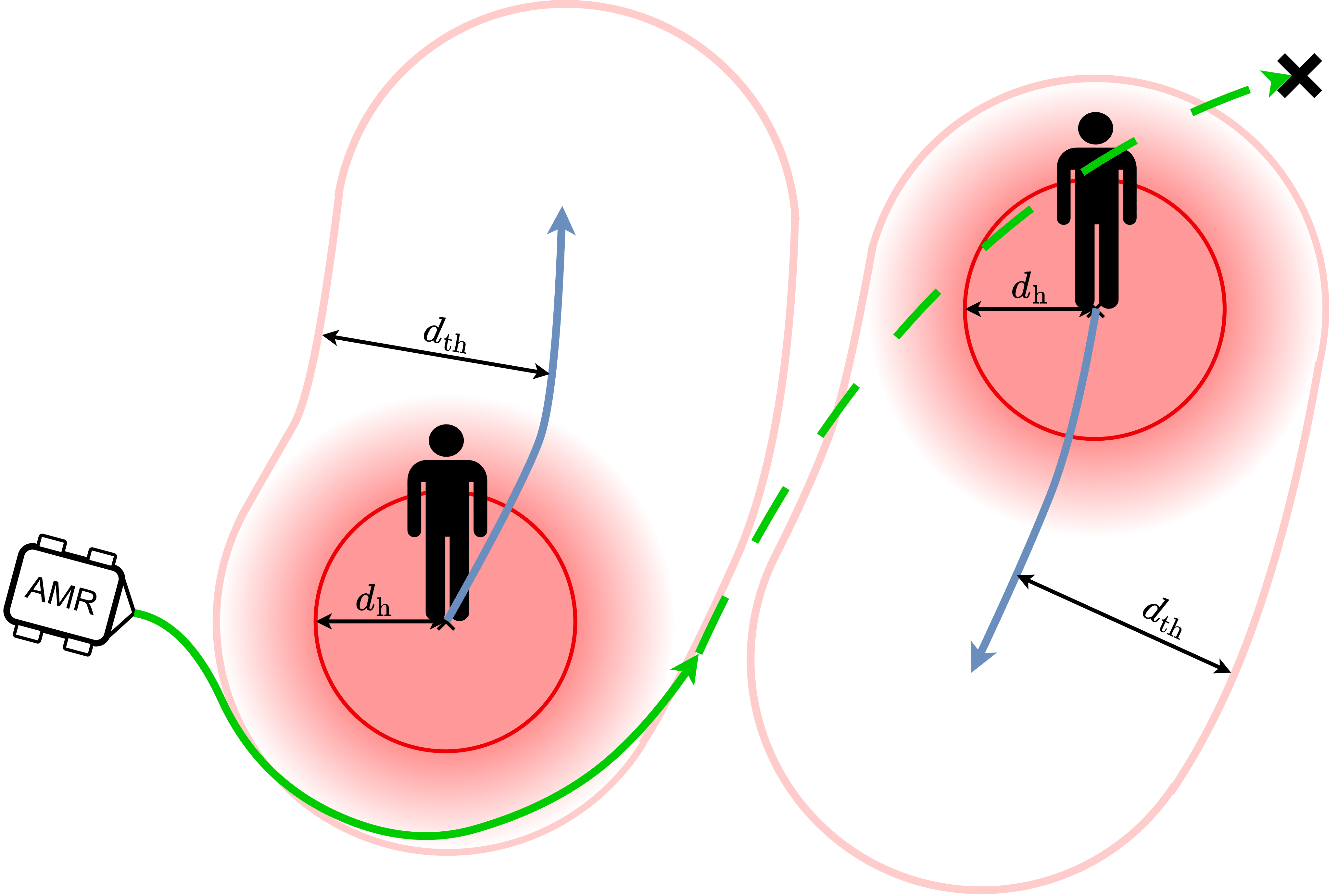}
            \caption{Human awareness considering areas for cost ($d_{\text{th}}$) and safety constraint ($d_\text{h}$) with the human predictions (\textbf{blue}) and the robots path (\textbf{green}).}
            \label{fig:human_awareness_graphic}
        \vspace{-5pt}
    \end{figure}
\begin{itemize}
    \item We propose an MPC formulation that includes a novel cost definition to achieve collision avoidance of nearby humans by considering their short-term predictions. The cost formulation implies a positive semi-definite Hessian that allows us to use recent fast embedded optimization tools, \emph{acados} \cite{Verschueren2021}, for achieving quasi-real-time motion generation (i.e., a few milliseconds for solving an MPC cycle).
    
    \item To further enhance its reliability, we interface the motion generation to feasibility monitors developed by considering established safety norms for mobile robotic applications.
\end{itemize}    
\section{RELATED WORK}
    First approaches to collision avoidance include model-based techniques not considering predictions of humans, e.g. Dynamic Window Approach (DWA) \cite{fox1997dynamic}, Elastic Band \cite{quinlan1993elastic} or Optimal Reciprocal Collision Avoidance (ORCA) \cite{van2008reciprocal}.
    
    Lately, there has been an increasing interest in MPC techniques for generating robot motion. Differently from \cite{fox1997dynamic,quinlan1993elastic, van2008reciprocal} they allow researchers to adopt more easy-to-use and customizable tools for numerical optimization.
    In this work, building on our previous results \cite{rudenkoIJRR2020, schoels2020ciao, schoels2020nmpc, heuerIROS2023}, we shift our attention towards the usage of fast embedded solvers for generating robot motion considering human motion predictions. 
    The idea of using a cost term in the MPC formulation to drive the trajectory away from dynamic obstacles has been explored in the past \cite{schaefer2021leveraging, kamel2017robust, schmerling2018multimodal}.
    In particular, several approaches have considered human motion predictions for generating robot motion \cite{schaefer2021leveraging, heuerIROS2023, schmerling2018multimodal,  chenCORL2020, nair2022stochastic}.
    Schmerling et al. \cite{schmerling2018multimodal} and Schaefer et al. \cite{schaefer2021leveraging} use multi-modal prediction in their MPC formulations, that include interactions between the agents in the scene. Heuer et al. \cite{heuerIROS2023} also use multi-modal predictions by directly considering the spatial relation between the predicted modes of motion.
    Our approach also uses predictions, differently, we adopt a formulation that is convex for all relevant values and thereby suited for fast embedded optimization using sequential quadratic programming (SQP) techniques \cite{diehl2002real}.
    
    Other approaches adopt data-driven technologies (i.e., deep reinforcement learning (DRL)) for the generation of robot motion among people: GA3C-CADRL \cite{everett2021collision} or SA-CADRL \cite{chen2017decentralized}. While being quite efficient in inferring robot actions, they often use a limited amount of predefined discrete actions due to complex training, and thus do not completely account for the dynamics of the robot and the environment leading to difficulties in ensuring safe behavior.
\section{BACKGROUND}
    \subsection{Safety Norms}\label{sec:background_safety}
        Existing safety standards must be taken into account when deploying novel planning and control algorithms to real robotic applications. Relevant in this regard is DIN EN ISO 3691-4 \cite{DIN_EN_3691} for driverless industrial trucks and DIN EN ISO 13482:2014-11 \cite{DIN_EN_ISO_13482} for personal care robots.
        The norms state that navigation controllers have to be designed to provide collision avoidance and risk reduction functionality. To this end, the controller has to constrain operational scenarios and the workspace around surrounding humans. 
        If there is a safety-relevant object, i.e., a human, in close proximity a protective stop is a sufficient action.
        \begin{remark}
            In this work, the term ``safety'' is regarded as compliant with the relevant safety norms \cite{DIN_EN_3691} and \cite{DIN_EN_ISO_13482}.
        \end{remark}
    \subsection{Human Motion Prediction}\label{subsec:HMP}
        As discussed in our previous work \cite{rudenkoIJRR2020}, the literature does not provide a consistent definition of motion predictions.
        \begin{remark}
        In this work, we assume the availability of all $i \in [1, N_{\text{h}}]$ humans positions $\mathbf{h}_{i, n}$ at time instant $n$ up to the prediction horizon $N$, i.e., predictions are reported as discrete trajectories $\mathbf{T}_{i, 1:N} = \{ \mathbf{h}_{i, 1}, \mathbf{h}_{i, 2}, \dots, \mathbf{h}_{i, N} \}$.
        \end{remark}
    \subsection{Fast Embedded Optimization with acados}
        We adopt \emph{acados} \cite{Verschueren2021}, a framework (i.e., a collection of solvers) for fast embedded optimal control (i.e., solving optimization problems with limited computational resources). Based on BLASFEO \cite{Frison2018blasfeo}, which is a library specifically designed for high-performance linear algebra techniques for embedded optimization, it offers several computationally efficient algorithms and components written in the \verb|C| programming language.
        \begin{remark}
        In our approach, we will use the SQP Real-Time Iteration (RTI) scheme \cite{diehl2002real} and the HPIPM solver with partial condensing \cite{frison2020hpipm}.        
        With the RTI scheme subsequent optimization problems are initialized and aligned to previous solutions, thus leading to reduced computation times \cite{diehl2002real}. In order to achieve fast computations and generate feasible solutions with SQP-RTI, we need to ensure that the cost-function derived Hessian stays positive semi-definite \cite{Verschueren2021}.        
        \end{remark}    
\section{APPROACH}
    Our approach consists of a combination of a novel human-aware collision avoidance formulation for fast embedded MPC together with safety-feasibility monitors. We describe the novel MPC formulation, the cost function, and the monitors in Sec.~\ref{sec:fastmpc} and \ref{sec:human_awareness}.    
    \subsection{Fast Embedded MPC Formulation}
    \label{sec:fastmpc}
        We construct the motion planner as a model predictive controller that solves a discretized optimal control problem (OCP) at each iteration.
        The OCP has the following structure:
        \vspace{-7pt}
        \begin{mini!}            
            {\mathbf{x}, \mathbf{u}}{\sum\limits_{n=0}^{N-1}
                    \biggl(\mathcal{J}_{\text{S}}(\mathbf{x}_n, \mathbf{u}_n, \mathbf{p}_n)\biggr) 
                    + \mathcal{J}_{\text{T}}(\mathbf{x}_N, \mathbf{p}_N) \label{eq:total_ocp_objective}}
            {\label{eq:total_ocp}}{}
            \addConstraint{\mathbf{x}_n}{\in \mathbb{X} \quad && n \in [0, N] \label{eq:total_ocp_1}}
            \addConstraint{\mathbf{u}_n}{\in \mathbb{U} \quad && n \in [0, N-1] \label{eq:total_ocp_2}}
            \addConstraint{\text{d}(\mathbf{x}_n, \mathbf{h}_{i, 0})}{\geq d_{\text{h}} \quad && n \in [0, N];\; i \in [1, N_{\text{h}}] \label{eq:total_ocp_3}}
            \addConstraint{\text{d}(\mathbf{x}_n, \mathbf{o})}{\geq d_{\text{s}} \quad && n \in [0, N] \label{eq:total_ocp_4}}
        \end{mini!}
        where $\mathbf{x}_n$ and $\mathbf{u}_n$ denote respectively the state and the control at a specific time step $n$ on the horizon which is split into $N$ shooting nodes, with $\mathbb{X}$ and $\mathbb{U}$ being the allowed state and control spaces. The additional parameters $\mathbf{p}_n$ include information on the goal $\mathbf{g}_n$, the predicted position of the human $\mathbf{h}_{\cdot, n}$ for all regarded $N_{\text{h}}$ humans and the position of the current closest static obstacle $\mathbf{o}$. The distance function $\text{d}(\cdot, \cdot)$ calculates the Euclidean distance. $d_{\text{h}}$ is the minimum safe distance from a human, and $d_{\text{s}}$ the allowed distance to static obstacles.
        The cumulative stage cost is comprised of:
        \begin{equation}
            \begin{split}
                \mathcal{J}_{\text{S}}(\mathbf{x}_n, \mathbf{u}_n, \mathbf{p}_n) = & \mathcal{J}_{\text{g}}(\mathbf{x}_n, \mathbf{p}_n) \\
                &+ \mathcal{J}_{\text{u}}(\mathbf{u}_n)\\
                &+ \mathcal{J}_{\text{col}}(\mathbf{x}_n, \mathbf{p}_n)~.
            \end{split}
        \end{equation}
        For the final shooting node, the terminal cost is:
        \begin{equation}
            \begin{split}
                \mathcal{J}_{\text{T}}(\mathbf{x}_n, \mathbf{p}_n) = & \mathcal{J}_{\text{g}}(\mathbf{x}_n, \mathbf{p}_n) \\
                &+ \mathcal{J}_{\text{col}}(\mathbf{x}_n, \mathbf{p}_n)~.
            \end{split}
        \end{equation}
        The goal cost term penalizes the distance to the given goal 
        \begin{equation}
            \label{eq:total_ocp_goal}
            \mathcal{J}_{\text{g}}(\mathbf{x}_n, \mathbf{p}_n) = {\left\lVert \mathbf{x}_n-\mathbf{p}_n \right\rVert}^{2}_{W_\text{g}} \qquad n \in [0, N]
        \end{equation}
        weighted by the diagonal matrix $W_\text{g}$.\\        
        The control is penalized with the control cost term
        \begin{equation}
            \label{eq:total_ocp_control}
            \mathcal{J}_{\text{u}}(\mathbf{u}_n) = {\left\lVert \mathbf{u}_n \right\rVert}^{2}_{W_\text{u}} \qquad n \in [0, N-1]
        \end{equation}
        weighted by the diagonal matrix $W_\text{u}$.
        The collision cost term $\mathcal{J}_{\text{col}}(\mathbf{x}_n, \mathbf{p}_n)$ is described in Sec.~\ref{sec:human_awareness}.
        
        The optimization is constrained by various inequality constraints.        
        General collision avoidance is achieved with \autoref{eq:total_ocp_4}. Aligned with the state of the art \cite{heuerIROS2023} this is implemented as a soft constraint that limits the reachable space around the position of the closest detected obstacle by not allowing the distance between the robot and the obstacle to be less than the distance $d_{\text{s}}$.
        Since the cost formulation does not provide guarantees the OCP has to be extended by a hard constraint to assure the safety of the surrounding humans.
        The state space is constrained by calculating the distance between the robot and the current position of the human for all surrounding humans and requiring this distance to be greater than a minimal distance $d_{\text{h}}$ which respects uncertainties of both the human perception system and the robot localization system in order to provide guarantees \cite{DIN_EN_ISO_13482}.
    \subsection{Human Awareness and Feasibility} \label{sec:human_awareness}
        \subsubsection{Human Awareness}
            General human awareness is achieved by using motion prediction trajectories of the humans in the robot's surroundings, i.e., $\mathbf{T}_{\cdot, 1:N}$. 
            
            To handle this complexity we propose a cost term based on a logistic function generating a potential field \cite{kamel2017robust}. However, the original term in \cite{kamel2017robust} is not $C^2$ (i.e., twice continuously differentiable) with a positive semi-definite Hessian and therefore not suitable for the efficient and reliable usage of SQP-RTI with \emph{acados} for fast embedded optimization since feasibility can not be guaranteed with a negative definite Hessian.
            Therefore, we adopt the following cost function:
            \begin{equation}
                \label{eq:convex_cost_term}
                \mathcal{J}_{\text{col}}(\mathbf{x}_n, \mathbf{p}_n) = \sum\limits_{i=1}^{N_{\text{h}}} \text{f}(\text{d}(\mathbf{x}_n, \mathbf{h}_{i, n}))
            \end{equation}
            \begin{equation}
                \label{eq:f}
                \text{f}(\text{d}(\mathbf{x}_n, \mathbf{h}_{i, n}))=
                \begin{cases}
                    \!\begin{aligned}
                        & \left(-\dfrac{\kappa q }{4}\right) \text{d}(\mathbf{x}_n, \mathbf{h}_{i, n}) \\
                        & + \left(\dfrac{q}{2}+\dfrac{\kappa q}{4} d_{\text{th}}\right)
                    \end{aligned}           & \text{if } \text{d}(\mathbf{x}_n, \mathbf{h}_{i, n}) \leq d_{\text{th}} \\[1ex]
                    \noalign{\vskip10pt}
                    \dfrac{q}{1 + e^{\kappa (\text{d}(\mathbf{x}_n, \mathbf{h}_{i, n}) - d_{\text{th}})}} & \text{if } \text{d}(\mathbf{x}_n, \mathbf{h}_{i, n}) > d_{\text{th}}
                \end{cases}
            \end{equation}
            Tuning of this cost function is possible by adjusting the smoothness with $\kappa > 0$ and by changing the threshold distance $d_{\text{th}}$ where the produced cost is $\dfrac{q}{2}$ which influences the distance the robot tries to keep from surrounding humans.        
            By using this cost function we aim to find optimal robot motion that maximizes the distance to humans considering their motion predictions (see \autoref{fig:human_awareness_graphic}).        
            The novel cost definition is convex for all mathematically relevant and reachable values, and therefore the requirement for the optimization calculation is met (i.e., the optimizer is pushed to find feasible solutions). 
            The linear function lays greater stress on the penalization of close distances between the robot and surrounding humans compared to a logistic potential field.\\
        \subsubsection{Feasibility and Safety Monitors}       
            We implement monitors that make sure that the controls sent to the robot actuators do not harm humans. 
            The safety system has to react to two critical possible situations that can occur during operation:
            \begin{enumerate}
                \item[C1)] The solver does not return a solution in time for real-time decision-making (e.g., overly complex formulation, numerical sub-optimality, infeasibility).
                \item[C2)] The solver does not find a safe solution. This can occur if there are numerical issues or calculations end up in constrained state space (e.g., a human is too close).                       
            \end{enumerate}
            Regarding case C1) we implement a running time check for the underlying solver. If a solution is not feasible and the optimization exceeds the allowed time frame provided by the real-time control loop appropriate measures can be initiated.
            To catch case C2) we use the optimization constraint based on the position of all surrounding humans (see  \autoref{eq:total_ocp_3} and \autoref{fig:human_awareness_graphic}). Regarding the relevant safety norms \cite{DIN_EN_3691} and \cite{DIN_EN_ISO_13482} this is sufficient as long as the source is reliable which is to be assumed and an adequate safety margin is included.
            Following the safety norms \cite{DIN_EN_3691} and \cite{DIN_EN_ISO_13482} the system commands a protective stop in both cases.
    \subsection{System Integration} \label{sec:system_integration}
        To facilitate easy integration across several robot platforms, we implement \texttt{HuMAN-MPC} into the Robot Operating System (ROS). The algorithm is embedded into a plugin for the local planner (ROS 2: Controller; ROS: move\_base). 
        Required for operation are two additional parts. One is a node responsible for the formulation of the OCP and the export of the library which is interfaced by \emph{acados} during runtime. The other one is an interface to connect the plugin to the prediction data of the surrounding humans.
\section{EXPERIMENTS}
    We design a set of experiments for evaluating \texttt{HuMAN-MPC}. Most of the experiments use Gazebo to simulate the environment and a TurtleBot3 with the ROS integration of \texttt{HuMAN-MPC} (see \autoref{fig:path_visualization}). The specifications of the PC used for the experiments are: Intel® Core™ i7-10850H, 32 GB RAM, NVIDIA Quadro RTX 3000.
    \subsection{Scenarios, Metrics, and Experiments}
        We design two main classes of experiments: one focused on the computation capabilities of the approach and the other on the analysis of the robot's behavior performance. While computational performance is the main aspect of the presented approach, latter experiments are conducted to show that the adaptions made to the formulation for fast embedded optimization include all the known advantages that come with using predictions.
        \paragraph{For computation efficiency}
            We design an experiment to show that \texttt{HuMAN-MPC} is efficient, even when considering a large number of surrounding humans and thereby a large number of optimization parameters.
            In this experiment, an increasing number of up to 30 randomly simulated humans are considered in the optimization.
        \paragraph{For behavior performance}
            We consider two scenarios:
            The \emph{random crowded and cluttered} scenario simulates ten human actors with random start and goal poses, which act according to the social force model, and 5  randomly placed static obstacles.
            In the \emph{crossing group} scenario the robot has to pass a group of nine simulated humans (i.e., actors) walking back and forth through the environment.
            To show advantages that arise from using predictions we also run the above-described scenarios to compare against the baseline: the open-source local planner Dynamic Window approach (DWB) \cite{dwb_2023}.
            Predictions are provided for the length of the optimization horizon using the constant velocity model \cite{scholler2020constant}.
    \subsection{Parameters and Model}
        In the experiments, we set the following values for the defined parameters: prediction horizon $5.0~\text{s}$, number of shooting nodes $50$,  stage goal cost weights $[0.5, 0.5, 0.0, 250.0]$,  terminal goal cost weights $[40.0, 40.0, 2.0, 0.0]$, control cost weights $[0.0, 0.0]$, collision cost: $q~=~2.0$ ($d_{\text{th}}~=~1$), $\kappa$~=~$5.0$. Constraint distances $d_\text{s}=~0.5~\text{m}$, $d_\text{h}=~0.5~\text{m}$ (including safety margin). 
        The model is a differential drive robot \cite{palmieri2014novel}.
\section{Results}
    \subsection{Real-Time Performance and Scalability}
        We observe from our results (mean iteration times are given in \autoref{tab:eval_iteration_time_scalability}) that the optimization steps on average do not take more than 5 ms, thus being well within the control frequency limits and thereby theoretically allowing for control frequencies even over 100 Hz.
        Furthermore, the embedded optimization scales notably well when increasing the number of parameters. \autoref{tab:eval_iteration_time_scalability}
        \begin{table}[h]
        \vspace{-7.5pt}
            \centering
            \begin{tabular}{cc|c|}
            \cline{3-3}
                                                                                                                                                               &                                                      & \begin{tabular}[c]{@{}c@{}}Mean iteration time {[}ms{]}\end{tabular} \\ \hline
            \multicolumn{1}{|c|}{\multirow{4}{*}{\begin{tabular}[c]{@{}c@{}} Num. of\\ humans -- (Num. of\\ parameters per\\ shooting node)\\\end{tabular}}} & \begin{tabular}[c]{@{}c@{}}5 -- (27)\end{tabular}   & 2.47                                                                     \\ \cline{2-3} 
            \multicolumn{1}{|c|}{}                                                                                                                             & \begin{tabular}[c]{@{}c@{}}10 -- (47)\end{tabular}  & 3.08                                                                     \\ \cline{2-3} 
            \multicolumn{1}{|c|}{}                                                                                                                             & \begin{tabular}[c]{@{}c@{}}20 -- (87)\end{tabular}  & 3.92                                                                     \\ \cline{2-3} 
            \multicolumn{1}{|c|}{}                                                                                                                             & \begin{tabular}[c]{@{}c@{}}30 -- (127)\end{tabular} & 4.75                                                                     \\ \hline
            \end{tabular}
            \caption{Iteration times considering a differing number of simulated humans.}
            \label{tab:eval_iteration_time_scalability}
            \vspace{-10pt}
        \end{table}
        
        Referring to the comparison experiments carried out in \cite{Verschueren2021} (see \autoref{tab:acados_comp_times_comparison}) it is furthermore shown that the \emph{acados} based optimization is superior in terms of computational efficiency (i.e., two orders of magnitude) to other relevant optimization solvers.        
        Aligned to this iteration times are much higher in our previous works \cite{heuerIROS2023}, \cite{schoels2020nmpc} and \cite{schoels2020ciao} with $34~\text{ms}$, $17.26~\text{ms}$ and $64~\text{ms}$ respectively. Note that the settings are not identical but only similar. Nevertheless, this still shows great progress in computational performance.
        \begin{table}[h]
                \vspace{-7.5pt}
                \centering
                \begin{tabular}{cc|c|c|c|c|}
                \cline{3-6}
                                                                                                                     &      & acados   & IPOPT    & ACADO    & GRAMPC   \\ \hline
                \multicolumn{1}{|c|}{\multirow{3}{*}{\begin{tabular}[c]{@{}c@{}}Iteration\\ time {[}ms{]}\end{tabular}}} & Mean & {1.05}    & 59.84    & 1.97     & 1.06     \\ \cline{2-6} 
                \multicolumn{1}{|c|}{}                                                                               & Min. & 0.87     & 49.06    & 1.90     & 0.81     \\ \cline{2-6} 
                \multicolumn{1}{|c|}{}                                                                               & Max. & 2.23     & 384.90   & 3.45     & 1.31     \\ \hline \hline
                \multicolumn{2}{|c|}{RCSO}                                                                                  & {1.01e-04} & 0.00e+00 & 1.01e-04 & 7.17e-02 \\ \hline
                \end{tabular}
                \caption{Case study results for \emph{acados} obtained in \cite{Verschueren2021}. \emph{acados} achieves the best mean MPC iteration time, with only minor relative cumulative sub-optimality (RCSO). IPOPT achieves the optimal solution while being on average 60 times slower than \emph{acados}.}
                \label{tab:acados_comp_times_comparison}
                \vspace{-15pt}
            \end{table}
        \begin{figure*}[t!]
        \vspace{-15pt}
            \centering
            \begin{subfigure}[c]{0.3\textwidth}
            \vspace{-15pt}
                 \centering
                 \includegraphics[width=\textwidth]{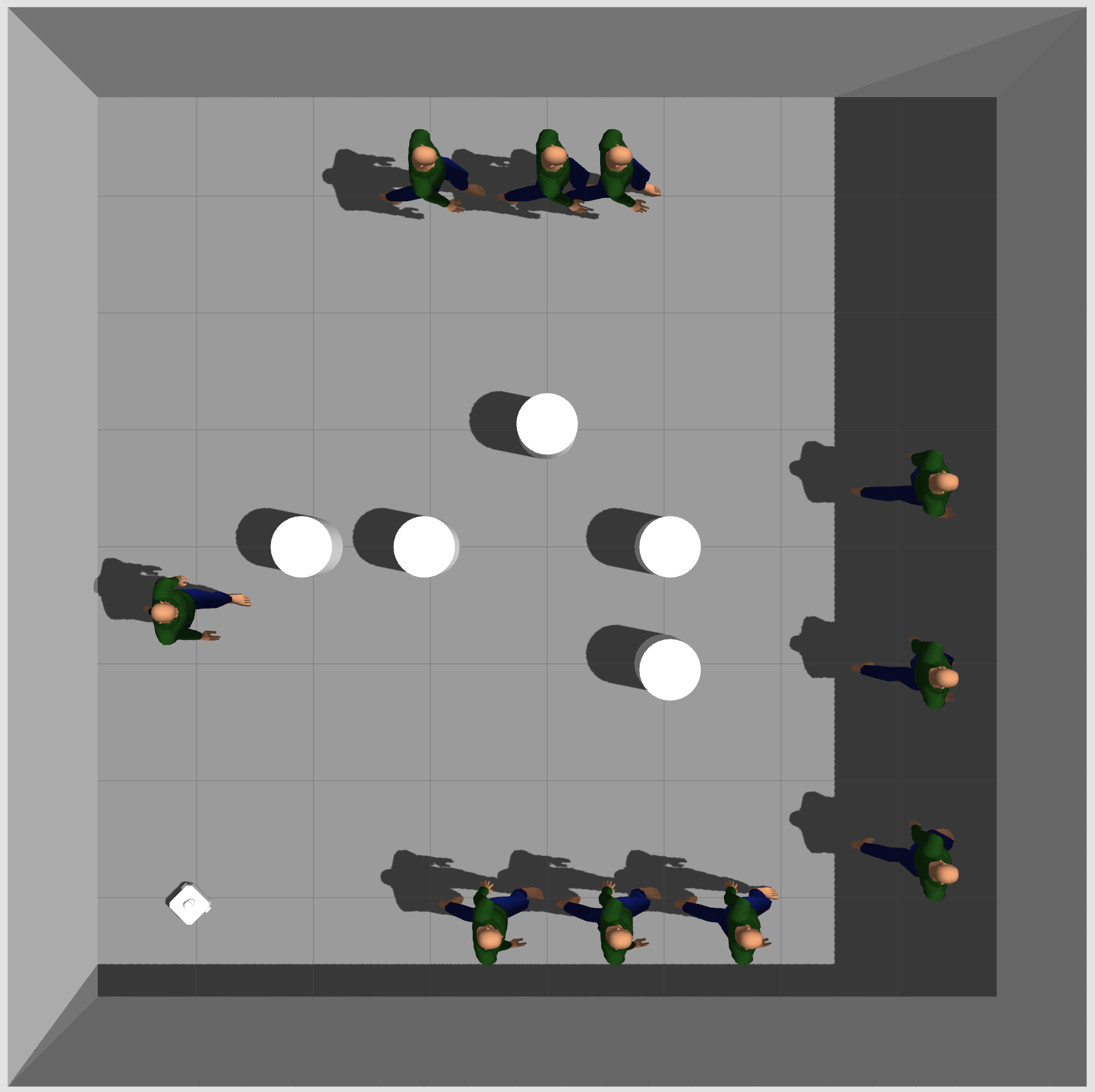}
                 \label{fig:crossing_group_sparse}
                 \vspace{-15pt}
            \end{subfigure}            
            \begin{subfigure}[c]{0.3\textwidth}
            \vspace{-15pt}
                 \centering
                 \includegraphics[width=\textwidth]{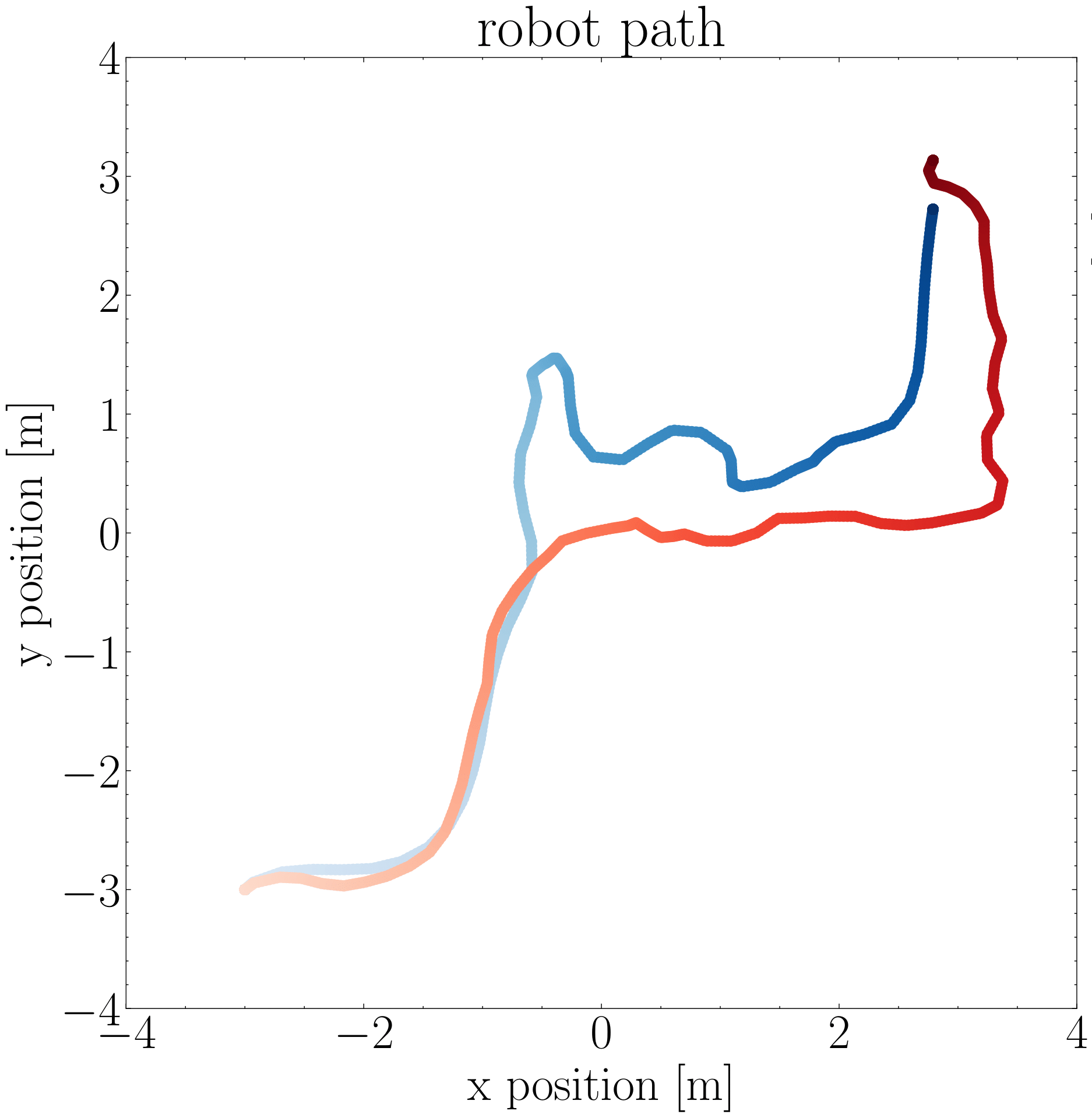}
                 \label{fig:random_crowded_cluttered}
                 \vspace{-15pt}
            \end{subfigure}
            \begin{subfigure}[c]{0.3\textwidth}
            \vspace{-15pt}
                 \centering
                 \includegraphics[width=\textwidth]{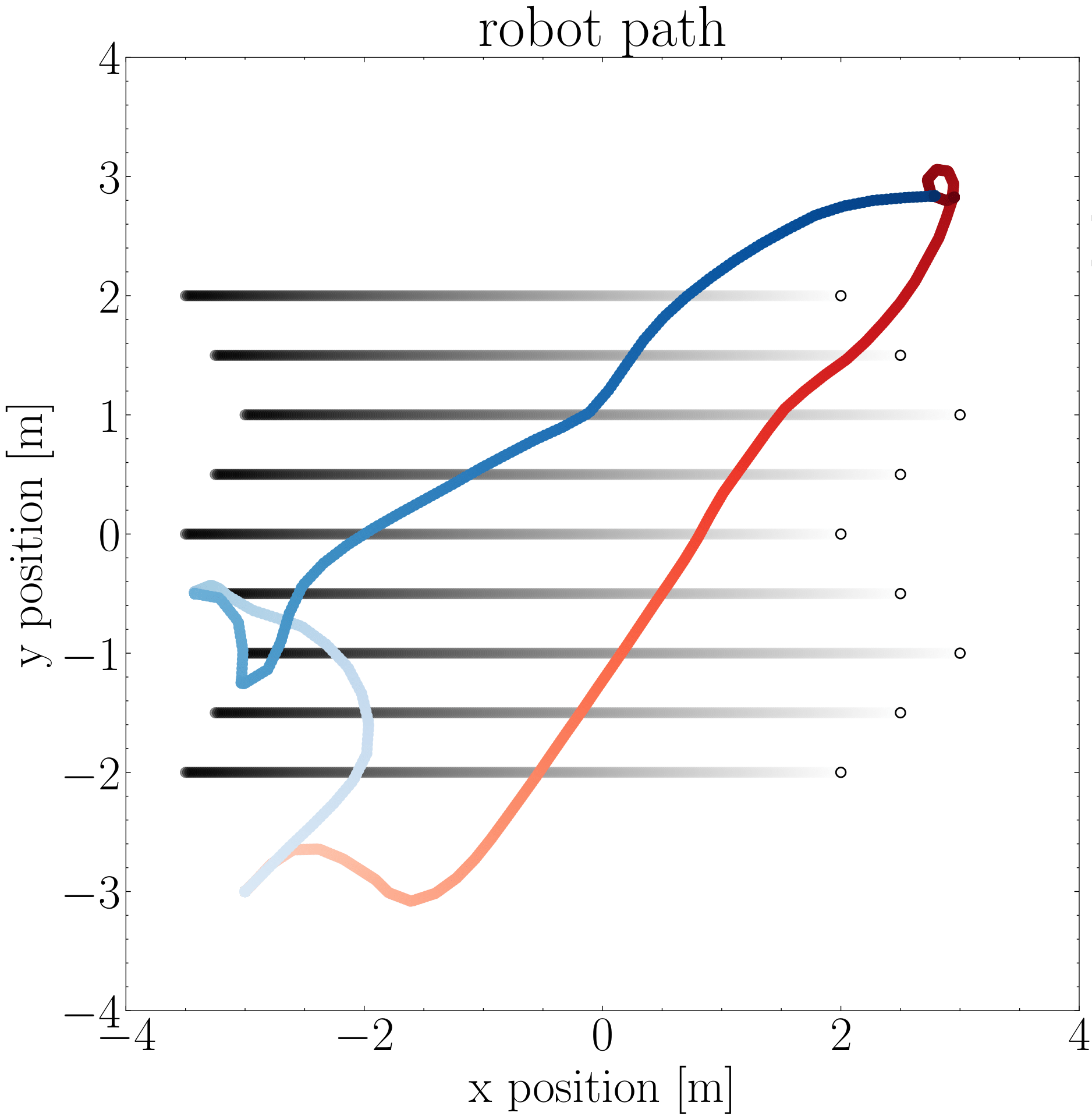}
                 \label{fig:crossing_group_dense}
                 \vspace{-15pt}
            \end{subfigure}            
            \caption{Left: Scene of the simulation environment with robot, human actors (start poses) and static obstacles; Center: Paths in random crowded and cluttered scenario; Right: Paths in crossing group scenario (\textbf{red}: \texttt{HuMAN-MPC}, \textbf{blue}: DWB)}
            \label{fig:path_visualization}
            \vspace{-15pt}
        \end{figure*}
    \subsection{Robot Behavior Performance}
        Evaluating both scenarios shows that making use of predictions allows \texttt{HuMAN-MPC} to navigate the environment more intuitively (i.e., passing behind the group, giving space to the humans), smoother and more legible (i.e., not getting stuck, not coming too close to humans). The resulting paths are shown in \autoref{fig:path_visualization}.
    \subsection{Feasibility and Safety Monitors}
        The monitoring system initiated a protective stop at every harmful occasion and at every occasion where optimization running times exceeded the limit.
        This goes to show that the implemented monitors (described in \autoref{sec:human_awareness}) are able to handle critical scenarios in compliance with the relevant safety norms \cite{DIN_EN_3691} and \cite{DIN_EN_ISO_13482}. 
        Furthermore, it is to note that our approach triggered the safety monitors less often for reasons of faster computation and development of fewer safety critical situations.
    \subsection{Real Robot}
\begin{figure}[h!]
            \centering
            \begin{subfigure}[c]{0.2\textwidth}
                \includegraphics[width=\textwidth]{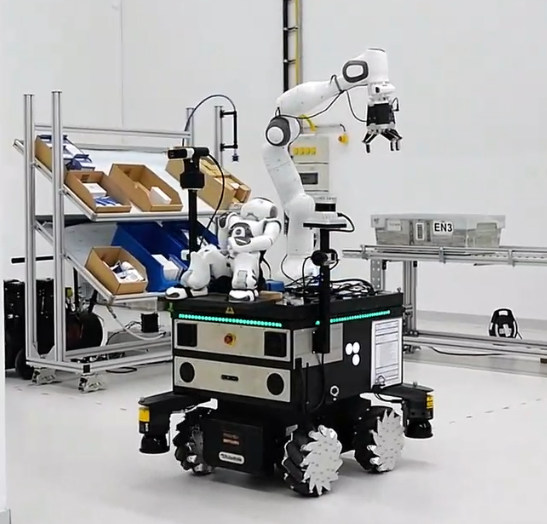}
            \end{subfigure}
            \begin{subfigure}[c]{0.24\textwidth}
                \includegraphics[width=\textwidth]{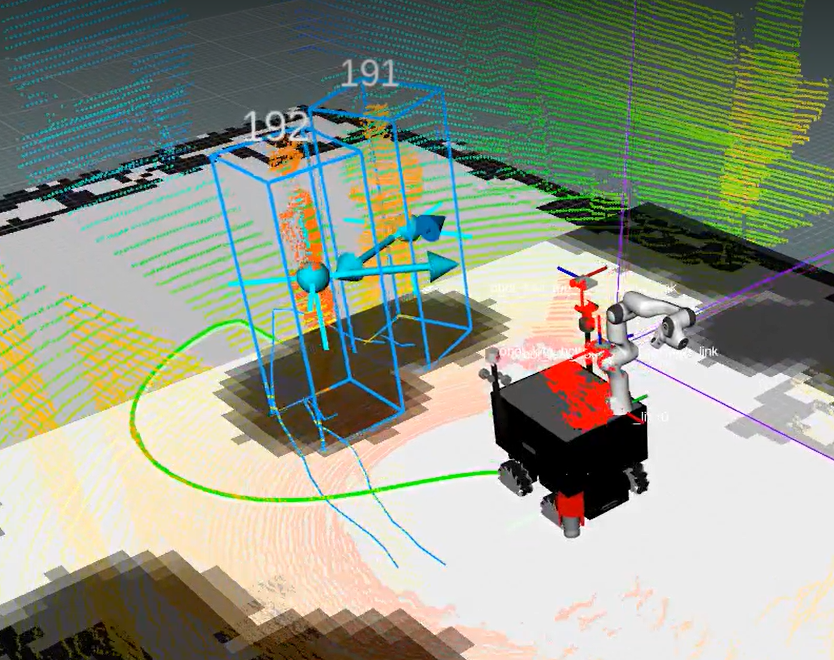}
            \end{subfigure}
            \caption{Left: DARKO robot. Right: Generated path.}
            \label{fig:DARKO}
            \vspace{-10pt}
        \end{figure}

         Additionally, we show functionality on a real robot using the robot platform of the research project DARKO\footnotemark \footnotetext{EU Project DARKO, \url{https://darko-project.eu/}}. The approach has been tested in several experiments which resulted in smooth movements without collisions. This also shows the advantages of the ROS integration. See \autoref{fig:DARKO}.
\section{CONCLUSIONS}
    To achieve efficient robot motion planning considering human motion predictions we propose \texttt{HuMAN-MPC}. The approach presents a novel formulation for fast embedded model predictive control. The optimization-based motion planning incorporates predictions of surrounding humans into an SQP-RTI scheme, while additionally verifying reliable and safe operation with monitors for feasibility and safety. Simulated and real-world experiments show that our approach is able to achieve substantially faster computation compared to other state-of-the-art baselines and provide great scalability while keeping predictive avoidance behavior.
    In future work, we plan to use more contextual information about the surroundings (e.g., semantics, human activities) leveraging the great scalability.

\footnotesize
\bibliographystyle{IEEEtran}
\bibliography{hielscherICRA2024}

\begin{thebibliography}{10}
\providecommand{\url}[1]{#1}
\csname url@samestyle\endcsname
\providecommand{\newblock}{\relax}
\providecommand{\bibinfo}[2]{#2}
\providecommand{\BIBentrySTDinterwordspacing}{\spaceskip=0pt\relax}
\providecommand{\BIBentryALTinterwordstretchfactor}{4}
\providecommand{\BIBentryALTinterwordspacing}{\spaceskip=\fontdimen2\font plus
\BIBentryALTinterwordstretchfactor\fontdimen3\font minus \fontdimen4\font\relax}
\providecommand{\BIBforeignlanguage}[2]{{%
\expandafter\ifx\csname l@#1\endcsname\relax
\typeout{** WARNING: IEEEtran.bst: No hyphenation pattern has been}%
\typeout{** loaded for the language `#1'. Using the pattern for}%
\typeout{** the default language instead.}%
\else
\language=\csname l@#1\endcsname
\fi
#2}}
\providecommand{\BIBdecl}{\relax}
\BIBdecl

\bibitem{rudenkoIJRR2020}
A.~Rudenko, L.~Palmieri, M.~Herman, K.~M. Kitani, D.~M. Gavrila, and K.~O. Arras, ``Human motion trajectory prediction: a survey,'' \emph{Int.~Journal of Robotics Research}, vol.~39, no.~8, 2020.

\bibitem{schoels2020ciao}
T.~Schoels, P.~Rutquist, L.~Palmieri, A.~Zanelli, K.~O. Arras, and M.~Diehl, ``Ciao⁎: Mpc-based safe motion planning in predictable dynamic environments,'' \emph{IFAC-PapersOnLine}, vol.~53, no.~2, 2020.

\bibitem{heuerIROS2023}
L.~Heuer, L.~Palmieri, A.~Rudenko, A.~Mannucci, M.~Magnusson, and K.~O. Arras, ``Proactive model predictive control with multi-modal human motion prediction in cluttered dynamic environments,'' in \emph{Int.~Conf.~Intell. Robot. Sys. (IROS)}.\hskip 1em plus 0.5em minus 0.4em\relax IEEE, 2023.

\bibitem{schoels2020nmpc}
T.~Schoels, L.~Palmieri, K.~O. Arras, and M.~Diehl, ``An {NMPC} approach using convex inner approximations for online motion planning with guaranteed collision avoidance,'' in \emph{Int.~Conf.~Robot. \& Autom. (ICRA)}.\hskip 1em plus 0.5em minus 0.4em\relax IEEE, 2020.

\bibitem{schaefer2021leveraging}
S.~Schaefer, K.~Leung, B.~Ivanovic, and M.~Pavone, ``Leveraging neural network gradients within trajectory optimization for proactive human-robot interactions,'' in \emph{Int.~Conf.~Robot. \& Autom. (ICRA)}.\hskip 1em plus 0.5em minus 0.4em\relax IEEE, 2021.

\bibitem{Verschueren2021}
\BIBentryALTinterwordspacing
R.~Verschueren, G.~Frison, D.~Kouzoupis, J.~Frey, N.~van Duijkeren, A.~Zanelli, B.~Novoselnik, T.~Albin, R.~Quirynen, and M.~Diehl, ``acados -- a modular open-source framework for fast embedded optimal control,'' \emph{Mathematical Programming Computation}, Oct 2021. [Online]. Available: \url{https://doi.org/10.1007/s12532-021-00208-8}
\BIBentrySTDinterwordspacing

\bibitem{fox1997dynamic}
D.~Fox, W.~Burgard, and S.~Thrun, ``The dynamic window approach to collision avoidance,'' \emph{IEEE Robot. \& Autom. Mag.}, vol.~4, no.~1, pp. 23--33, 1997.

\bibitem{quinlan1993elastic}
S.~Quinlan and O.~Khatib, ``Elastic bands: Connecting path planning and control,'' in \emph{Int.~Conf.~Robot. \& Autom. (ICRA)}.\hskip 1em plus 0.5em minus 0.4em\relax IEEE, 1993.

\bibitem{van2008reciprocal}
J.~Van~den Berg, M.~Lin, and D.~Manocha, ``Reciprocal velocity obstacles for real-time multi-agent navigation,'' in \emph{Int.~Conf.~Robot. \& Autom. (ICRA)}.\hskip 1em plus 0.5em minus 0.4em\relax IEEE, 2008.

\bibitem{kamel2017robust}
M.~Kamel, J.~Alonso-Mora, R.~Siegwart, and J.~Nieto, ``Robust collision avoidance for multiple micro aerial vehicles using nonlinear model predictive control,'' in \emph{IEEE/RSJ Int. Conf. Intell. Robot. Sys. (IROS)}, 2017, pp. 236--243.

\bibitem{schmerling2018multimodal}
E.~Schmerling, K.~Leung, W.~Vollprecht, and M.~Pavone, ``Multimodal probabilistic model-based planning for human-robot interaction,'' in \emph{Int.~Conf.~Robot. \& Autom. (ICRA)}.\hskip 1em plus 0.5em minus 0.4em\relax IEEE, 2018.

\bibitem{chenCORL2020}
Y.~Chen, U.~Rosolia, C.~Fan, A.~Ames, and R.~M. Murray, ``Reactive motion planning with probabilistic safety guarantees,'' in \emph{Conf. Robot Learn. (CoRL)}, vol. 155, 2020, pp. 1958--1970.

\bibitem{nair2022stochastic}
S.~H. Nair, V.~Govindarajan, T.~Lin, C.~Meissen, H.~E. Tseng, and F.~Borrelli, ``Stochastic mpc with multi-modal predictions for traffic intersections,'' in \emph{2022 IEEE 25th Int. Conf. on Intelligent Transportation Systems (ITSC)}, 2022, pp. 635--640.

\bibitem{diehl2002real}
M.~Diehl, H.~G. Bock, J.~P. Schl{\"o}der, R.~Findeisen, Z.~Nagy, and F.~Allg{\"o}wer, ``Real-time optimization and nonlinear model predictive control of processes governed by differential-algebraic equations,'' \emph{J.~Process Control}, vol.~12, no.~4, pp. 577--585, 2002.

\bibitem{everett2021collision}
M.~Everett, Y.~F. Chen, and J.~P. How, ``Collision avoidance in pedestrian-rich environments with deep reinforcement learning,'' \emph{IEEE Access}, vol.~9, pp. 10\,357--10\,377, 2021.

\bibitem{chen2017decentralized}
Y.~F. Chen, M.~Liu, M.~Everett, and J.~P. How, ``Decentralized non-communicating multiagent collision avoidance with deep reinforcement learning,'' in \emph{Int.~Conf.~Robot. \& Autom. (ICRA)}.\hskip 1em plus 0.5em minus 0.4em\relax IEEE, 2017.

\bibitem{DIN_EN_3691}
``\textsc{DIN EN ISO 3691-4:2020-11}; industrial trucks - safety requirements and verification - part 4: Driverless industrial trucks and their systems (\textsc{ISO} 3691-4:2020),'' Nov. 2020.

\bibitem{DIN_EN_ISO_13482}
``\textsc{DIN EN ISO 13482:2014-11}; robots and robotic devices - safety requirements for personal care robots (\textsc{ISO} 13482:2014),'' Nov. 2014.

\bibitem{Frison2018blasfeo}
G.~Frison, D.~Kouzoupis, T.~Sartor, A.~Zanelli, and M.~Diehl, ``Blasfeo: basic linear algebra subroutines for embedded optimization,'' \emph{{ACM} Transactions on Mathematical Software}, vol.~44, no.~4, pp. 1--30, 2018.

\bibitem{frison2020hpipm}
G.~Frison and M.~Diehl, ``Hpipm: a high-performance quadratic programming framework for model predictive control,'' \emph{IFAC-PapersOnLine}, vol.~53, no.~2, pp. 6563--6569, 2020.

\bibitem{dwb_2023}
\BIBentryALTinterwordspacing
{DWB} controller. [Online]. Available: \url{https://github.com/ros-planning/navigation2/tree/main/nav2_dwb_controller}
\BIBentrySTDinterwordspacing

\bibitem{scholler2020constant}
C.~Sch{\"o}ller, V.~Aravantinos, F.~Lay, and A.~Knoll, ``What the constant velocity model can teach us about pedestrian motion prediction,'' \emph{IEEE Robotics and Automation Letters}, vol.~5, no.~2, pp. 1696--1703, 2020.

\bibitem{palmieri2014novel}
L.~Palmieri and K.~O. Arras, ``A novel {RRT} extend function for efficient and smooth mobile robot motion planning,'' in \emph{2014 IEEE/RSJ Int. Conf. on Intelligent Robots and Systems}, 2014.

\end{thebibliography}
\end{document}